# Py-Feat: Python Facial Expression Analysis Toolbox

*Jin Hyun Cheong[1*], Eshin Jolly[1*†], Tiankang Xie[1,2*], Sophie Byrne[1], Matthew Kenney[1], & Luke J. Chang[1†]*

[1] Computational Social and Affective Neuroscience Laboratory
Department of Psychological & Brain Sciences
Dartmouth College
Hanover, NH 03755

[2] Department of Quantitative Biomedical Sciences
Geisel School of Medicine
Dartmouth College
Hanover, NH 03755

[*]Equal contributions
[†]Corresponding author: eshin.jolly@dartmouth.edu, luke.j.chang@dartmouth.edu

## Abstract

Studying facial expressions is a notoriously difficult endeavor. Recent advances in the field of affective computing have yielded impressive progress in automatically detecting facial expressions from pictures and videos. However, much of this work has yet to be widely disseminated in social science domains such as psychology. Current state of the art models require considerable domain expertise that is not traditionally incorporated into social science training programs. Furthermore, there is a notable absence of user-friendly and open-source software that provides a comprehensive set of tools and functions that support facial expression research. In this paper, we introduce Py-Feat, an open-source Python toolbox that provides support for detecting, preprocessing, analyzing, and visualizing facial expression data. Py-Feat makes it easy for domain experts to disseminate and benchmark computer vision models and also for end users to quickly process, analyze, and visualize face expression data. We hope this platform will facilitate increased use of facial expression data in human behavior research.

## Introduction

Facial expressions can reveal insights into an individual's internal mental state and provide nonverbal channels to aid in interpersonal and cross-species communication [1,2]. One of the main challenges to studying facial expressions has been arriving at a consensus understanding as to how to best represent and objectively measure expressions. The Facial Affect Coding System (FACS) [3] is one of the most popular systems to reliably [4] quantify the intensity of groups of facial muscles referred to as action units (AUs). However, extracting facial expression information using FACS coding can be a laborious and time-intensive process. Becoming a certified FACS coder requires 100 hours of training, and manual labeling is slow (e.g., one minute of video can take an hour [5]) and inherently contains cultural biases and errors [6,7]. Facial electromyography (EMG) provides one method to objectively record from a finite number of facial muscles at a high temporal resolution [8,9], but requires specialized recording equipment that restricts data collection to the laboratory and can visually obscure the face making it less ideal for social contexts.

Automated methods using techniques from computer vision have emerged as a promising approach to extract representations of facial expressions from pictures, videos, and depth cameras both inside and outside the laboratory. Participants can be untethered from cumbersome wires and can naturally engage in tasks such as watching a movie or having a conversation [10–14]. In addition to AUs, computer vision techniques have provided alternative embedding spaces to represent facial expressions such as facial landmarks [15] or lower dimensional latent representations [16]. These tools have a number of applications relevant to psychology such as predicting the intensity of emotions [17–20] and other affective states such as pain [21,22], distinguishing between genuine and fake expressions [23], detecting signs of depression [24], inferring traits such as personality [25–27] or political orientations [28], and predicting the development of interpersonal relationships [12,14]. Though facial expression research has seen rapid growth in affective computing facilitated by recent advances in machine learning, adoption in fields outside the domain of computer science such as psychology has been surprisingly slow.

In our view, there are at least two specific barriers contributing to the slow adoption of automated methods in social science fields such as psychology. First, there is a relatively high barrier to entry to training and accessing state of the art models capable of quantifying facial expressions. This requires knowledge of computer vision techniques, neural network architectures, and access to large labeled datasets and computational infrastructure that include Graphics Processing Units (GPUs). Though there are impressive efforts to share high quality datasets [29–35], there are still difficulties sharing this data involving participants' privacy, complicated end user agreements, expensive handling fees, contacting data curators, and finding affordable and stable long-term hosting solutions. Though hundreds of models have been developed to characterize facial expressions, no standards have emerged for disseminating these models to end users. These models are typically reported in conference proceedings, occasionally shared on open code repositories such as Github, and require considerable domain knowledge as they have been developed using a multitude of computer languages, rarely have documentation, and occasionally have restrictive licensing. Each model may require the data to be preprocessed in a specific way or rely on additional features (e.g., landmarks, predefined regions of interest). Because there are currently no generally agreed upon standards for training and benchmarking beyond data competitions (e.g., WIDER, 300W, FERA, etc), each model is typically trained on different datasets, which makes it difficult to benchmark the models using the same dataset to aid in the model selection process [17,36]. Platforms such as paperswithcode.com are helping to standardize the dissemination and benchmarking of models, but sharing state of the art models has not yet become a norm in the field. Other domains such as natural language processing and reinforcement learning have begun to overcome this issue with a variety of high quality software platforms such as Stanza [37], SpaCy, OpenAI Gym [38], and HuggingFace.

Second, there is a notable lack of free open-source software to aid in detecting, preprocessing, analyzing, and visualizing facial expressions (Table 1). Commercial software options such as Affdex ([Affectiva Inc](...)) available through iMotions [39] and Noldus FaceReader [40] can be expensive, have limited functionality, and typically do not employ state of the art models [41–43] (see [17,20] for commercial software performance comparisons). Furthermore, due to strong interest from industry, there have been several free software packages such as the Computer Expression Recognition Toolbox [44], Intraface [15], and Affectiva API [45] ([Affectiva Inc](...)) that have turned into commercial products or been acquired by larger technology companies such as Apple Inc or Meta and rendered unavailable to researchers. Currently, OpenFace [46] is the most widely used open-source software that allows users to extract facial landmarks and action units from face images and videos. However, OpenFace does not provide a comprehensive suite of tools for preprocessing, analyzing, and visualizing data, which would make these tools more accessible to non-domain experts. As an example, in other fields such as neuroscience, the rapid growth of neuroimaging research has been facilitated by the widespread use of free tools such as FSL [47], AFNI [48], SPM [49], and NiLearn [50] that enables end users to preprocess, analyze, and visualize complex brain imaging data. We believe the broader emotion research community would greatly benefit from additional software platforms dedicated to facial expression analysis with functions for extracting, preprocessing, analyzing, and visualizing facial expression data.

|  | Facial feature detection | | | | | Preprocessing | Analysis | Free |
|---|---|---|---|---|---|---|---|---|
|  | **Facial landmarks** | **Action units** | **Emotions** | **Headpose** | **Gaze** | | | |
| iMotions* | | | | | | X | X | |
| FACET | X | X | X | X | | | | |
| AFFDEX | X | X | X | X | | | | |
| Noldus FaceReader | | X** | X | | | X | X** | |
| OpenFace | X | X | | X | X | *** | | X |
| face-api.js | X | | X | | | | | X |
| Py-Feat | X | X | X | X | | X | X | X |

*Table 1. Software comparison on functionalities and affordability. X indicates features provided by each package. Features from Py-Feat toolbox are shown in brackets. Facial landmarks are points pertaining to locations of key spatial positions of the face including the jaw, mouth, nose, eyes, and eyebrows. Action units are facial muscle groups defined by FACS [51]. Emotions refer to the detection of canonical emotional expressions. Headpose refers to the pitch, roll, and yaw orientations of the face. Gaze refers to the direction the eyes are looking. \*iMotions is a platform and its feature extraction relies on the purchase of either the AFFDEX or FACET modules. \*\*Detection of action units and analysis functionalities require a separate add-on purchase of The Action Unit Module and the Project Analysis Module for the Noldus FaceReader. \*\*\*We note that OpenFace can perform some preprocessing such as median face image subtraction and post-processing of AUs to correct for at-rest expressions.*

To meet this need, we have created the Python Facial Expression Analysis Toolbox (Py-Feat) which is a free, open-source package dedicated to support the analysis of facial expression data. It provides tools to extract facial features like OpenFace [46], but additionally provides modules for preprocessing, analyzing, and visualizing facial expression data (Figure 1). Py-Feat is designed to meet the needs of two distinct types of users. Py-Feat benefits computer vision researchers who can use our platform to disseminate their state of the art models to a broader audience and easily compare their models with others on the same benchmark metrics. It also benefits social science researchers looking for free and easy to use tools that can both detect and analyze facial expressions. In this paper, we outline the key components of the Py-Feat toolbox including the facial feature detection module and analysis tools, provide quantitative assessments of the performance of the detection models on benchmark data including the robustness of the models to real world data, and provide a tutorial of how the toolbox can be used to analyze an open face expression dataset.

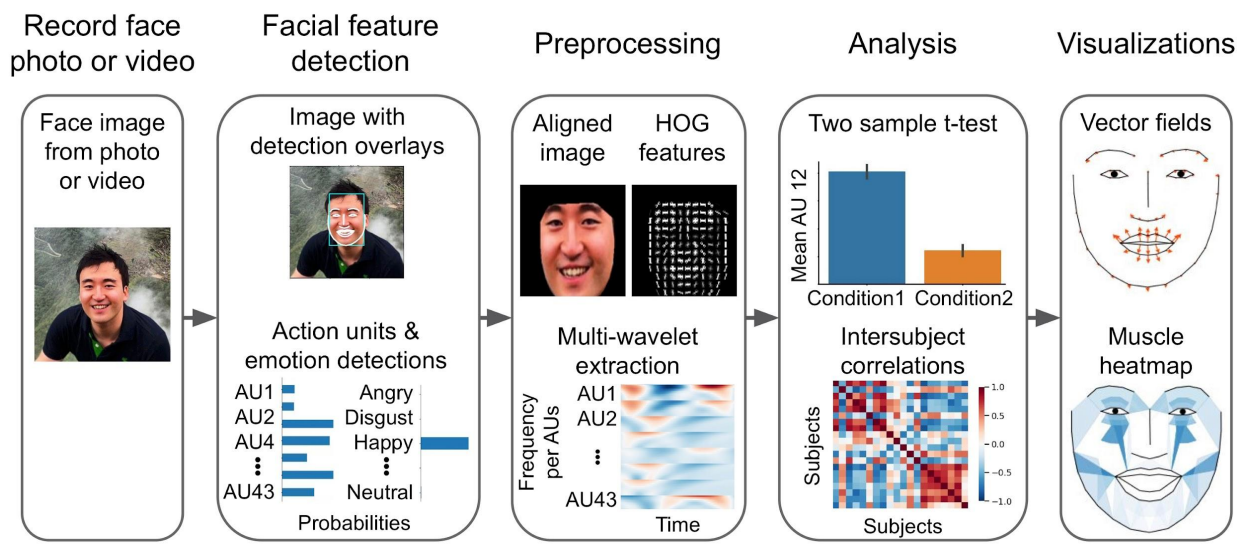

*Figure 1. Facial expressions analysis pipeline. Analysis of facial expressions begins with recording face photos or videos using a recording device such as webcams, camcorders, head mounted cameras, or 360 cameras. After capturing the face, researchers can use Py-Feat to detect facial features such as the location of the face within a rectangular bounding box, the location of key facial landmarks, action units, and emotions, and check the detection results with image overlays and bar graphs. The detection results can be preprocessed by extracting additional features such as Histogram of Oriented Gradients (HOG) or multi-wavelet decomposition. Resulting data can then be analyzed within the toolbox using statistical methods such as t-tests, regressions, and intersubject correlations. Visualization functions can generate face images from models of action unit activations to show vector fields depicting landmark movements and heatmaps of facial muscle activations.*

## Py-Feat Design and Module Overview

Py-Feat is written in the Python programming language. We selected Python over other popular languages (e.g., Matlab, C, etc) for several reasons. First, Python is open source and completely free to use and compiles to all major operating systems (e.g., Mac, Windows, Unix). This makes the software accessible to the largest number of users. Second, Python is among the easiest programming languages to read and learn and is increasingly being taught in introduction to programming classes. Though we do not currently provide a graphical user interface (GUI) to Py-Feat, we believe it is highly easy to use with minimal background in programming (see our example code below). Third, Python has emerged as one of the primary languages used across academia and industry for data science. There is a vibrant developer community that has already created a rich library of tightly integrated high quality scientific computing packages for working with: arrays such as numpy [52] and pandas [53]; scientific numerical routines with scipy [54], machine learning algorithms with scikit-learn [55], tensorflow [56], and py-torch [57]; and plotting with matplotlib [58], seaborn [59], and plotly. This makes it easy for Py-Feat to incorporate new functionality as it becomes available in other toolboxes, but also for Py-feat users to incorporate any Python package into other processing pipelines. Many of the core libraries are supported by big tech companies and are rapidly providing functionality to enable users to take advantage of newer innovations in hardware such as GPUs and distributed computing systems. In addition, Python libraries tend to have comprehensive documentation and testing and there are many excellent tutorials for learning how to use python online, which makes the language very accessible to beginners. For example, we have developed basic tutorials for learning to analyze data with Python on our DartBrains.org course [60] and more advanced tutorials on analyzing naturalistic neuroimaging data [61]. We have built a jupyter-book [62] to accompany our toolbox with tutorials on how to perform analyses that can be easily augmented by the user community (https://py-feat.org/).

Py-Feat currently has two main modules for working with facial expression data. First, the `Detector` module makes it easy for users to detect facial expression features from image or video stimuli. We offer multiple models for extracting the primary face expression features that most end users will want to work with. This includes detecting faces in the stimuli and identifying the coordinates of the spatial location of a bounding box for each face. We also detect 68 facial landmarks, which are coordinates identifying the spatial location of the eyes, nose, mouth, and jaw. The bounding box and landmarks can be used in models to detect the head pose such as the face orientation in terms of rotation around axes in three-dimensional space. Py-Feat also detects higher level facial expression features such as AUs and basic emotion categories. We offer multiple models for each detector to keep the toolbox flexible for many use cases, but we also have picked sensible defaults for users who may be overwhelmed by the number of options. The features cover the majority of the ways in which facial expressions can be currently described by computer vision algorithms. Importantly, new features and models can be added to the toolbox as they become available in the field. The majority of the models in the toolbox are implemented in PyTorch [57], which means they can also utilize Nvidia GPUs if they are available, which can dramatically speed up performance.

In addition, Py-feat also includes the `Fex` data module to work with the features extracted from the `Detector` module. This module includes methods for preprocessing, analyzing, and visualizing facial expression data. We offer an easy to use application programming interface (API) for slicing, grouping, sampling, and summarizing data as well as selecting different types of data (i.e., faceboxes, landmarks, action units, emotions, face poses), preprocessing facial expression time series data, extracting additional features from time series data, analyzing aggregates of facial expressions data, and visualizing intermediary preprocessing steps.

## Py-Feat Performance

Computer vision models are highly complex and often employ completely different preprocessing steps and model architectures. All of the technical details about the architecture of each of the models and how they were trained can be found in the Supplementary Materials. To provide users with an estimate of how well these models are likely to perform on their own datasets, we report benchmark performance on datasets that were never used in training the models. Importantly, we primarily used benchmark datasets that are the standard for each domain in data competitions and include highly variable naturalistic images collected in the wild when possible. Table 2 includes details about each of the benchmark datasets. Full details can be found in the supplementary materials.

| Dataset Name | Benchmark Type | Participants | Images | Type of data | Posed | Annotations |
|---|---|---|---|---|---|---|
| WIDER | Face Bounding Box | 393,703 | 32,203 | Images retrieved from search engines | In the wild | Manually annotated face location |
| 300W | Landmarks | >600 | 600 | Images retrieved from search engines | In the wild | Semi-Automatic & Manual Corrections |
| BIWI Kinect | Head Pose | 20 | 15,000 | Video recorded while subjects rotate their head | Posed | Semi-Automatic |
| DIFSA+ | Action Units | 9 | 57,000 | 3-minute video, imitate 30 facial action units | Spontaneous, Posed | Manually annotated AUs (1, 2, 4, 5, 6, 9, 12, 17, 20, 25, 26) by trained FACS coder |
| Namba | Action Units | 12 | 288 | Images taken at different angles | Posed | Manually annotated AUs by trained FACS coder |
| AffectNet | Emotions | 440,000 | 440,000 | Images retrieved from search engines | In the wild | Manually annotated emotion categories (neutral, surprise, happy, fear, sad, disgust, contempt, anger) |

*Table 2. Benchmarking datasets. Details about each dataset used for benchmarking the py-feat detectors.*

*Face detection*

One of the most basic steps in the facial feature detection process is to identify if there is a face in the image and where that face is located. Py-Feat includes three popular face detectors including Faceboxes [63], Multi-task Convolutional Neural Network (MTCNN) [64,65], and RetinaFace [66]. These detectors are widely used in other open-source software [46] and are known to achieve fast and accurate face detection results even for partially occluded or non-frontal faces. Face detection results are reported as a rectangular bounding box of the face and includes a confidence score for each detected face. We benchmarked the face detection models on the validation set of the WIDER FACE dataset, which is a standard dataset containing images in the wild retrieved from the internet [67], using average precision described in the WIDER Face technical paper [68]. Overall, we found that the Py-Feat implementations of each of the models achieved acceptable levels of performance, although lower than what was reported in the original papers [66] (Table 3). This may be a consequence of using different hyperparameters. We also observed decreased performance as the classification task becomes increasingly more difficult, which includes small, inverted, and highly occluded faces.

| **Model** | **Easy** | **Medium** | **Hard** |
|---|---|---|---|
| Feat-img2Pose constrained | 0.589 | 0.576 | 0.351 |
| Feat-img2Pose unconstrained | 0.740 | **0.744** | **0.555** |
| Feat-Faceboxes | 0.537 | 0.348 | 0.147 |
| Feat-MTCNN | 0.725 | 0.718 | 0.473 |
| Feat-RetinaFace (default) | [0.760] | [0.669] | [0.347] |

*Table 3. Benchmarking results for face bounding box detection. Easy, Medium, Hard results retrieved from WIDER Face. Numbers are average precision scores with higher numbers indicating better detection accuracy. Bold numbers indicate best performance for each column and bracketed numbers indicate the performance of the model selected as the default for Py-Feat.*

*Landmark detection*

After a face is identified in an image, it is common to identify the facial landmarks, which are coordinate points in the image space outlining the jaw, mouth, nose, eyes, and eyebrows of a face. The distance and angular relationships between the landmarks can be used to represent face expressions and used to infer affective states such as pain [21]. Py-feat uses a standard 68-coordinate facial landmark scheme that is widely used across datasets and software [46,69,70] and currently includes three facial landmark detectors including the Practical Facial Landmark Detector (PFLD) [71], MobileNets [72], and MobileFaceNets [73] algorithms. We benchmarked these models on the 300 Faces in the Wild (300W) dataset [70,74], which is a standard used in data competitions and contains in-the-wild face images that vary across luminance, scale, pose, expressions and occlusion levels. We compute the average root mean squared error between the predicted and ground truth coordinates across the landmark points normalized by the interocular distance. Overall, we found that the Feat-MobileFaceNet performed the best on our benchmark.

| Model | 300W-Test RMSE |
|---|---|
| Feat-MobileNet | 5.78 |
| Feat-MobileFaceNet (default) | **[4.99]** |
| Feat-PFLD | 5.39 |

*Table 3. Benchmarking results for face landmark detection. Feat models were initialized with face bounding boxes using RetinaFace. Numbers are root mean squared errors of coordinates with lower numbers indicating better alignment. Bolded numbers indicate best performance and bracketed numbers indicate the performance of the model selected as the default for Py-Feat.*

## Head pose detection

Another feature of a face expression beyond its location in an image or the location of specific parts of the face is the position of the head in three dimensional space. Rotations from a head on view can be described in terms of rotation around the x, y, and z planes and are referred to as pitch, roll, and yaw respectively. Py-feat includes support for the Img2Pose model. This model does not rely on prior face detections, so it can also be used as a face bounding box detector. The constrained version of Img2Pose is fine-tuned on the 300W-LP dataset, which only includes head poses in range (-90° to +90°). We benchmarked our head pose models using the BIWI Kinect dataset, which contains videos of participants rotating their heads according to specific pose instructions [75]. We computed the Mean Absolute Error in degrees for pitch, roll and yaw. Overall, we found that the constrained version of Img2Pose achieved a slightly better performance compared to the unconstrained version on our benchmark.

| Model | Pitch MAE | Roll MAE | Yaw MAE | Average MAE |
|---|---|---|---|---|
| Img2pose constrained | **[3.96]** | [4.74] | [3.65] | **[4.12]** |
| Img2pose unconstrained | 5.97 | **4.45** | **3.36** | 4.59 |

*Table 4: Model Performance on BIWI Kinect Head Pose Dataset. Model performance on the BIWI Kinect dataset, where Mean Absolute Error (MAE) values are reported in degrees (lower is better). Table shows performance of the img2pose models. Bolded numbers indicate best performance and bracketed numbers indicate the performance of the model selected as the default for Py-Feat.*

## Action unit detection

In addition to the basic properties of a face in an image, py-feat also includes models for detecting deviations of specific facial muscles (i.e., action units; AUs) from a neutral face expression using the FACS coding system. Py-feat currently contains two models for detecting action units. The architecture of the models are based on the highly robust and well-performing model used in OpenFace [46], which extracts Histogram of Oriented Gradient (HOG) features from within the landmark coordinates using a convex hull algorithm, compresses the HOG representation using Principal Components Analysis (PCA), and finally uses these features to individually predict each of the 12 AUs using popular shallow learning methods based on kernels (i.e., linear Support Vector Machine; SVM [76]), and ensemble learning (i.e., optimized gradient boosting; XGB [77]) (see supplemental materials for training details). We compare the performance of our models to OpenFace and also FACET, which was previously available in iMotions before the company was acquired by Apple Inc. We benchmarked the AU detection models using the Extended DISFA Plus dataset [33], which contains short videos of participants making posed facial expressions based on imitating a target image and also spontaneous facial expressions elicited from viewing experimental stimuli. We used F1 scores, an accuracy metric for binary classification, to quantify the performance of twelve different AUs. We found that the previously available FACET-iMotions achieved the best overall accuracy and was the best detector for AUs 2, 4, 9, 15, and 17. OpenFace and our Feat-XGB model achieved the second highest average F1 scores followed by the Feat-SVM model. OpenFace was the most accurate in detecting AUs 1, 6, and 12. The Feat-XGB model performed the best on AUs 5, 20, and 25, while the Feat-SVM model only performed the best on AU26. We have selected the Feat-XGB model to be the default model as it provides AU detection probability estimates rather than binary classifications.

|  | Model | AU1 | AU2 | AU4 | AU5 | AU6 | AU9 | AU12 | AU15 | AU17 | AU20 | AU25 | AU26 | Average |
|---|---|---|---|---|---|---|---|---|---|---|---|---|---|---|
| Models not in Py-Feat | FACET iMotions | .58 | **.62** | **.74** | .56 | .78 | **.73** | .77 | **.59** | **.47** | .15 | .64 | .43 | **.59** |
|  | OpenFace | **.71** | .52 | .69 | .49 | **.81** | .54 | **.83** | .34 | .43 | .13 | .72 | .67 | .57 |
| Models in Py-Feat | Feat-XGB | [.60] | [.54] | [.64] | [**.57**] | [.71] | [.39] | [.78] | [.25] | [.26] | [**.35**] | [**.84**] | [.56] | [.54] |
|  | Feat-SVM | .50 | .46 | .73 | .46 | .68 | .59 | .78 | .20 | .29 | .19 | .74 | **.69** | .53 |

*Table 6. Benchmarking results for AU models on DisfaPlus. Numbers shown are F1 scores. Bolded numbers indicate best performance and bracketed numbers indicate the performance of the model selected as the default for Py-Feat.*

## *Emotion detection*

Finally, Py-feat also includes models for detecting the presence of specific emotion categories based on third party judgments. Emotion detectors are trained on manually posed or naturalistically elicited emotional facial expressions which allows detectors to classify new images based on how much a face resembles a canonical emotional facial expression. It is important to note that there is currently no consensus in the field if categorical representations of emotion are the most reliable and valid nosology of emotional facial expressions [78,79]. For example, detecting a smiling face as happy does not necessarily imply that the individual is experiencing an internal subjective state of happiness [80], as these types of latent state inferences require additional contextual information beyond a static image [81]. However, labeling specific configurations of AUs with the semantic concepts of emotions can still be useful in emotion research to characterize the contexts in which people tend to display these facial expressions or how the display of certain emotion expressions accompanies changes in learning [82] and social behaviors [14]. Py-feat includes two emotion detectors capable of detecting seven categories of emotions: anger, disgust, fear, happiness, sadness, surprise, and neutral. The Residual Masking Network (ResMaskNet) [83] is an end-to-end convolutional neural network model that combines deep residual networks with masking blocks. The masking blocks help focus the model's attention on local regions of interest to refine its feature map for more fine-grained predictions and the residual structure helps to maintain performances in deeper layers. We also provide a statistical learning model that uses Linear SVM [76] using a similar procedure as our AU models. We benchmarked our models using F1 scores on a random subset of 500 images from the AffectNet dataset [84], which contains unposed expressions of emotions as they naturally occur in the wild outside of a carefully curated laboratory environment. We found that the Residual Masking Network model [83] achieved the highest F1 score, followed by the FACET-iMotions model, and the statistical learning models trained on HOG features.

|  | Model | Anger | Disgust | Fear | Happy | Sad | Surprise | Neutral | Average |
|---|---|---|---|---|---|---|---|---|---|
| Models not available on Py-Feat | FACET iMotions | .33 | .42 | .35 | .67 | .24 | .36 | .43 | .40 |
| Models available on Py-Feat | Residual Masking Network (default) | **[.53]** | **[.53]** | **[.48]** | **[.77]** | **[.54]** | **[.55]** | **[.49]** | **[.55]** |
|  | Feat-SVM | .39 | .28 | .37 | .64 | .33 | .39 | .34 | .39 |

*Table 7. Benchmarking results for motion models on AffectNet. Numbers shown are F1 scores. Bolded numbers indicate best performance and bracketed numbers indicate the performance of the model selected as the default for Py-Feat.*

## Robustness Experiments

While computer vision researchers typically focus on developing new face expression models that can outperform previous work on standard benchmarking datasets, end users are often more interested in how well the models perform on real world data collection contexts. This type of data is typically messier than the carefully curated open datasets. We intentionally selected benchmark datasets that contain spontaneous or naturalistic images collected outside the laboratory in the wild. In addition to these benchmarks, we also evaluated the robustness of the models included in Py-feat to different types of real world scenarios that are known to create problems for computer vision models including variations in luminance, occlusions of specific regions of the face, and also head rotation.

### *Luminance*
To test the robustness of our model to different lighting conditions, we modified our benchmark datasets to include two different levels of luminance (low, where brightness factor uniformly sampled from [0.1, 0.8] for each image and high, where brightness factor uniformly sampled from [1.2, 1.9] for each image). This can be useful for knowing how the models might be impacted by inconsistent lighting or smaller variations in skin pigmentation. Overall, we found that the majority of the deep learning detectors were fairly robust to variations in luminance. However, the shallow learning detectors that rely on HOG features were more dramatically impacted by high and low levels of variance (Figure 2).

*Occlusion*

In addition, we evaluated the performance of all of the detectors in three different occlusion contexts. Occlusions of the face are very common in real world data collection scenarios where a participant may cover their face with a hand, or be partially hidden behind some other physical object. We separately masked out the eyes, nose, and mouth on the benchmark datasets described above by applying a black mask to regions of the face using the facial landmark information (Figure 2A). The pose and landmark models were fairly robust to facial occlusions. However, face detection substantially dropped with occlusions, particularly when the nose was masked. Occlusion of specific facial structures can also provide an interesting lesion test for higher level facial feature extraction such as action units and emotions. Consistent with our expectations, the AU detector performance dropped for AUs 1,2,4,5,6,9 when the eyes were masked, while performance dropped for AUs 12, 15, 20, 25,& 26 when the mouth was masked. AU 9 and 20 detection performance dropped when the nose was blocked. The emotion models were even more dramatically affected by occlusion of specific facial structures. Anger, fear, sadness, and surprise detection was substantially impacted by occlusion of the eyes, while disgust, happy, and neutral detection dropped when the mouth was blocked, and Anger, Disgust, Fear, and Sadness were degraded with occlusions to the nose.

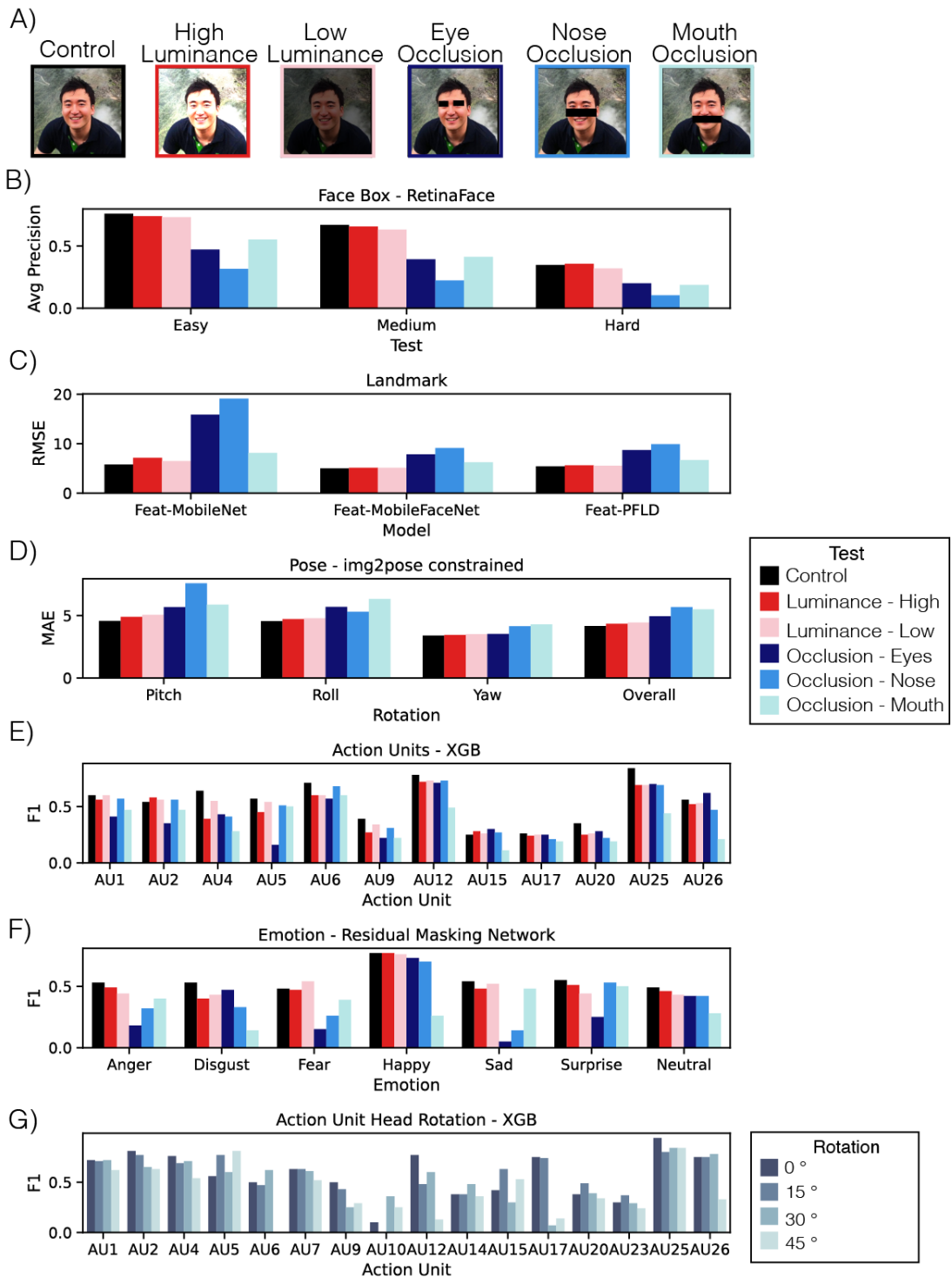

*Figure 2. Py-feat Detector Robustness Experiments.* A) Example image for robustness manipulations. B) RetinaFace face detection robustness results. Values are Average Precision where larger indicates better performance. C) Landmark detection robustness results. Values are Normalized Mean Average Error (MAE) where smaller values indicate better performance. D) img2pose-constrained pose detection robustness results. Values are Mean Average Error (MAE) where smaller values indicate better performance. E) Feat-XGB AU detection robustness results. Values are F1 scores where larger values indicate better performance. We note that the DISFA+ dataset does not include labels for AU7. F) Residual Masking Network emotion detection robustness results. Values are F1 scores

*where larger values indicate better performance. G) Feat-XGB AU robustness to rotation results. Values are F1 scores where larger values indicate better performance.*

## *Robustness against Head Rotation*

Most action unit models are trained using images in which the participants directly face the camera. However, in real world situations, faces are likely to be rotated relative to the camera position. Prior work has evaluated the performance of different AU detection algorithms on a new dataset, in which participants (N=12) were instructed to imitate specific facial expressions, while a camera recorded their expressions at specific rotation angles of 0°, 15°, 30° and 45° [85]. Action units for each image were manually annotated by a trained FACS coder. We tested our py-feat-XGB AU detection model using this dataset and found that AU detection performance tends to decrease as rotation angles increase. However, the XGB model is fairly robust to rotation for most of the AUs except for AUs 9, 12, 17, & 26, where performance drops substantially for the largest 45° rotation (Figure 2G).

## Visualization

We provide several plotting tools to help visualize the `Fex` detection results in each stage of the analysis pipeline. In the facial feature detection stage, we offer the `plot_detections` function that overlays the face, facial landmarks, action units, and emotion detection results in a single figure (Figure 1). This function can be used to validate the detection results at each video frame or image. The `Fex` class also allows users to plot time series graphs as well, which can be useful for examining how detected action unit activities vary over time or if there are segments of missing data.

In addition, we provide a model which can be used to visualize how combinations of activated AUs will look like on a stylized anonymous face Figure 3. This model visualizes the intensity of AUs overlaid onto a face in the approximate locations of where the facial muscles are located and also how AUs deform the face. Using this model, users can visualize the action units and their accompanying 2D landmark deformation on a standard face from any combination of action unit activations identified from their analyses (see supplemental materials for training details) [22,86]. We hope to incorporate other types of visualization models as they become available.

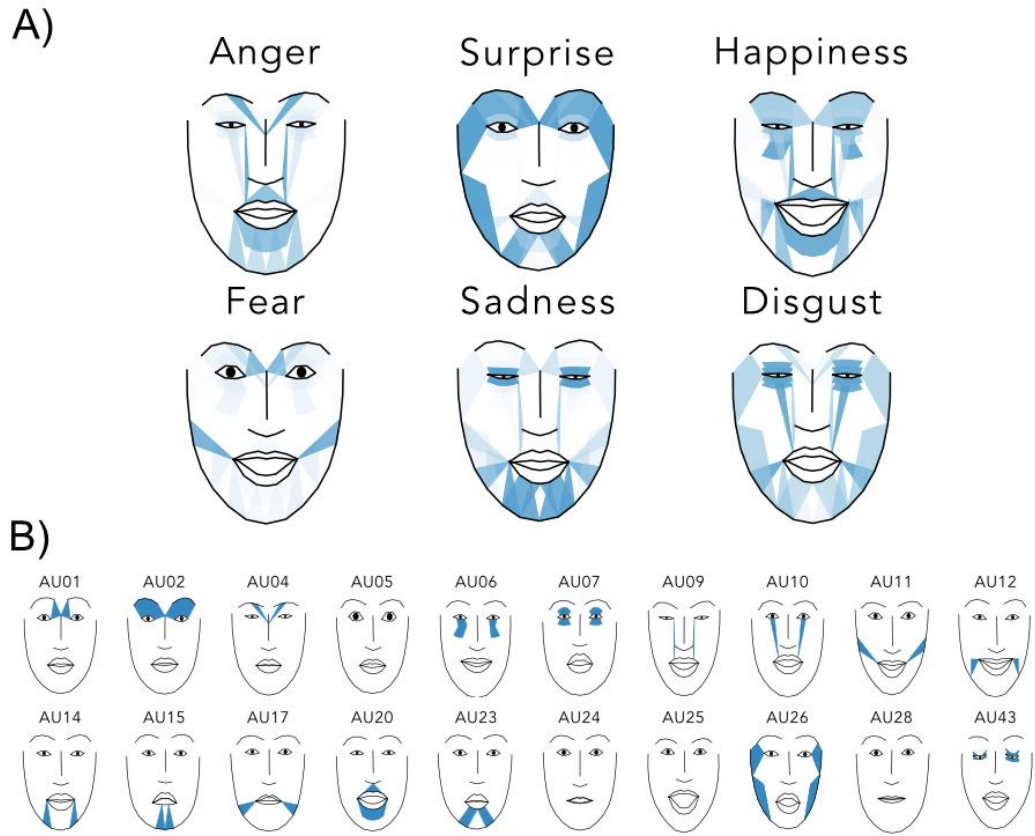

**Figure 3 | Demonstration of action unit to landmark visualization. (A):** Facial expressions generated from AU detections on real images. Detected AU activations were extracted from each of six labeled images displaying one emotion and projected through Py-Feat's visualization model. **(B):** Facial expressions generated by manually activating each AU in sequence.

## Example Py-feat Analysis Walkthrough

Py-feat easily facilitates numerous complex analyses. As a demonstration, we used a subset of the open video dataset from [87] in which participants were filmed while speaking in two conditions: delivering *good* news statements (e.g., "your application has been accepted" ) or *bad* news statements (e.g., "your application was denied"). A more comprehensive walkthrough using these data is included in the Py-Feat full analysis tutorial.

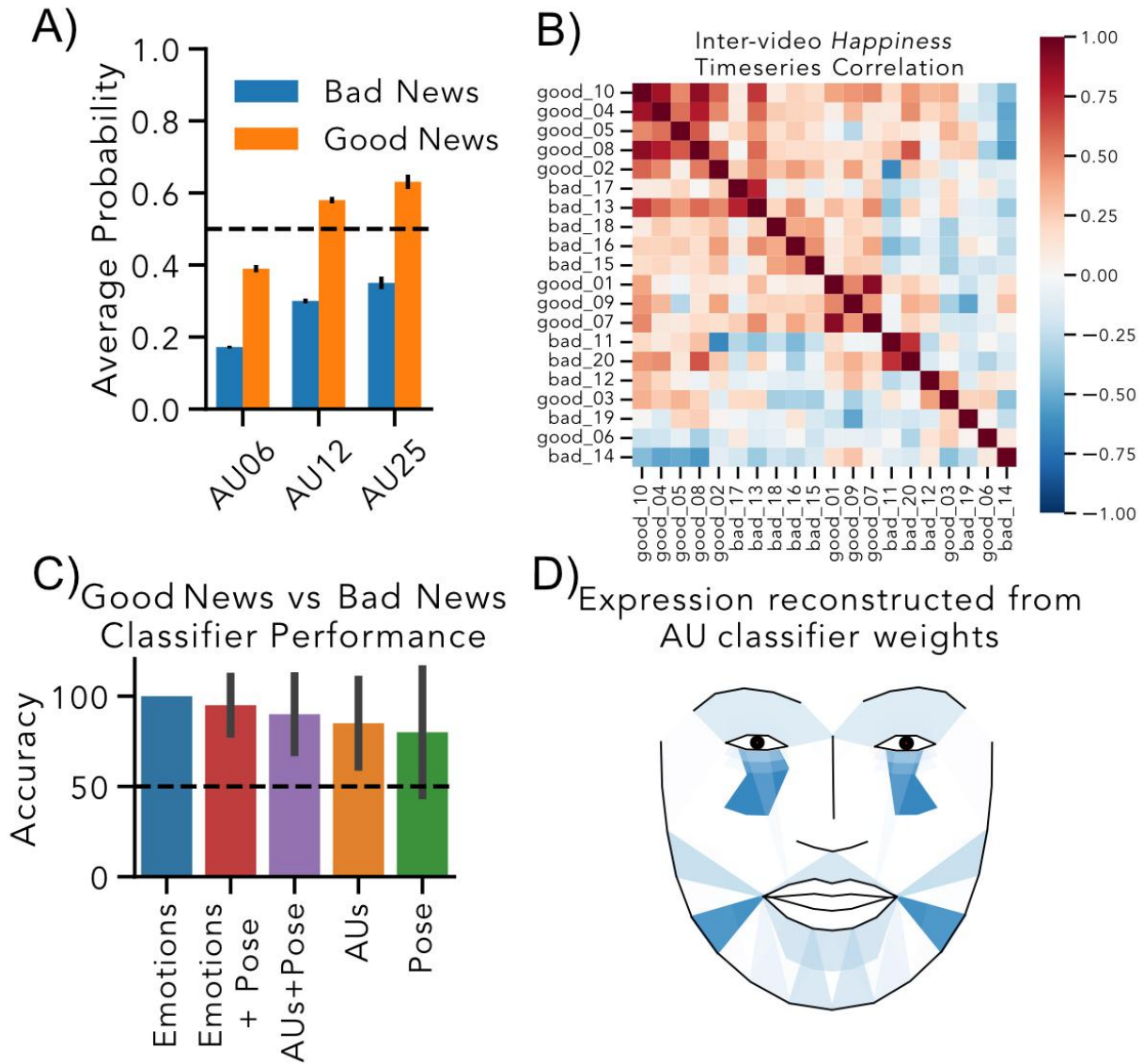

**Figure 4 | Illustrative Py-Feat Analyses. (A):** Average probability of action unit (AU) activation differences when delivering good news and bad news for AUs 6, 12, and 25. The dashed line reflects maximal detector uncertainty. **(B)**: Clustered intervideo time-series correlations of happiness detected over video-frames. Warmer colors indicate a pair of videos was more similar in terms of their Happiness time-courses. **(C)**: Example replication analysis of Watson et al [87]. Each bar depicts the cross-validated accuracy decoding good vs bad news clips using emotion, AU, pose or combined features. Error-bars reflect the standard deviation across cross-validation folds. Py-Feat's default emotion detector performs perfectly on the subset of data in this example. The dashed line reflects chance performance. **(D)**: Facial expression reconstructed from the AU classifier weights using the AU decoder (orange bar).

Extracting facial features can be extracted in py-feat with relative ease using an intuitive API, and only requires two lines of code: one to initialize a `Detector` and another to process a video:

```
detector = Detector() # initialize default detectors
fex = detector.detect_video('video.mp4') # process each video frame
```

The `fex` object is a dataframe organized as frames by features, and contains *all* detections for every frame of the video including: faceboxes, landmarks, poses, action units, and emotions. Each `fex` object makes use of a special `.sessions` property that facilitates easy data aggregation and comparison. For example, we can compare the means of each condition of the data by setting sessions to the condition labels with `.update_sessions()`, followed by `.extract_summary()` to compute summary statistics aggregated by condition (Fig 4A):

```
# dictionary mapping video name to the condition it belonged to
By_condition = fex.update_sessions(
      {'001': 'good_news', '002': 'bad_news', ...})

# plot condition mean per action unit
by_condition.extract_mean().aus.plot(kind="bar")
```

Py-feat also makes it easy to perform time series analyses using the `.isc()` method. For example, we can estimate the similarity between videos in terms of how their detected happiness varies over time (Fig 4B):

```
# calculate the pairwise similarity between videos in terms of their detected
happiness
intervideo_similarity = fex.isc(col = "happiness", method='pearson')

# visualize the video x video correlation matrix
from seaborn import heatmap
heatmap(intervideo_similarity)
```

Py-feat makes it simple to perform formal comparisons using the `.regress()` method. This method performs a "mass-univariate" style analysis [88] across all specified features. For example, we can use the experiment condition labels ("good" or "bad" news) as contrast codes and AUs as outcomes to perform a t-test on every AU. This returns the associated regression beta-values, standard-errors, t-statistics, p-values, degrees-of-freedom, and residuals for each AU:

```
# setup mean difference contrast of good news > bad news
by_condition_codes = fex.update_sessions({"goodNews": 1, "badNews": -1})

# compare condition differences at every AU
b, se, t, p, df, residuals = by_condition_codes.regress(
           X="sessions", y="aus", fit_intercept=True)
```

Py-feat can just as easily facilitate a decoding analysis like the classification analysis performed by Watson and colleagues [87] using the `.predict()` method (Fig 4C). For example, we can use all AUs as features and try to classify the condition in which participants' were delivering news. This returns the decoder model object along with its cross-validated performance:

```
# same algorithm used by Watson et al
from sklearn.discriminant_analysis import LinearDiscriminantAnalysis

# predict conditions from AUs with 5-fold cross-validation
model, cross_validated_accuracy = by_condition_codes.predict(
  X="aus", y="sessions", model=LinearDiscriminantAnalysis)
```

Unique to Py-Feat is the ability to use its visualization model to reconstruct any facial expression from AU values (Fig 4D). A compelling use case is reconstructing the facial expression implied by the weights estimated for each AU by the decoder. Py-feat offer two functions to do this: `plot_face()` which reconstructs a single image and `animate_face()` which can morph one facial expression to another to emphasize what is changing:

```
from feat.plotting import plot_face, animate_face

# static image of reconstructed face
plot_face(au=model.coef_.squeeze()) # the LDA classifier weights

# animated gif morphing a neutral face to the reconstructed face
animate_face(
 start="neutral", # start with a neutral facial expression
 end=model.coef_.squeeze() # end with LDA classifier weights
)
```

These simple examples are only a fraction of the analyses that are possible using py-feat, but provide an example of how the toolbox makes it possible to conduct complex analyses with minimal python code.

## Discussion

In this paper, we describe the motivation, design principles, and core functionality of the open-source Python package Py-Feat. This package aims to bridge the gap between model developers creating new algorithms for detecting faces, facial landmarks, action units, and emotions with end users hoping to use these cutting edge models in their research. To achieve this, we designed an easy to use and open-source Python toolbox that allows researchers to quickly detect facial expressions from face images and videos and subsequently preprocess, analyze, and visualize the results. We hope this project will make facial expression analysis more accessible to researchers who may not have sufficient domain knowledge to implement these techniques themselves. In addition, Py-Feat provides a platform for model developers to disseminate their models to end-user researchers and compare the performance of their model with others included in the toolbox.

PYTHON FACIAL EXPRESSION ANALYSIS TOOLBOX

Automated detection of facial expressions has the potential to complement other techniques such as psychophysiology, brain imaging, and self-report [14,22,89] along with 3-D simulations [90] in improving our understanding of how emotions interact with perception, cognition, and social interactions and are impacted by our physical and mental health. Studying facial expressions is becoming increasingly more accessible to non-specialists. For example, recording participants has become more convenient with a number of affordable recording options such as webcams that can be used to record remote participants, open-source head mounted cameras allowing reliable face recordings in social settings [13], as well as 360 cameras that can be used to record multiple individuals simultaneously. The primary goal of Py-Feat is to make the preprocessing, analysis, and visualization of these results similarly accessible and free of charge to non-specialists. Open source software focused on the full analysis pipeline has been instrumental in contributing to the rapid progress of research in other domains such as neuroimaging with FSL [47], AFNI [48], SPM [49], and NiLearn [50] and natural language processing with Stanza [37], SpaCy, and HuggingFace. We believe the broader emotion research community would greatly benefit from additional software platforms dedicated to facial expression analysis with functions for extracting, preprocessing, analyzing, and visualizing facial expression data.

Our toolbox is designed to be flexible and dynamic and includes models that are performing near state of the art. However, there are several limitations that are important to note. First, our current implementations of some of the models are not performing as well as the original versions. This could be attributed to nuances in hyperparameter optimization, variations in random seeds, and variations in the benchmarking datasets. We anticipate that these models will improve over time as more datasets become available and also plan to continually incorporate new models as they become available. Benchmarking of new models will be added to a living document on our project website to allow users to make informed choices in selecting models. Second, we have not yet attempted to optimize our toolbox for speed. For example, we did not benchmark our models on processing time because we believe most users will be applying these detectors on batches of pre-recorded videos rather than in real-time applications. Currently, our models are able to process a single image in about 400 milliseconds with a GPU and about 1.5 seconds on a CPU. For users who need faster processing times on videos, processing can be sped up by temporally downsampling and skipping frames. We hope to optimize our code and improve processing time in future versions of our toolbox. Third, our models likely contain some degree of bias with respect to gender and race. We have attempted to use as much high quality publicly available data as possible to train our models and selected challenging real world datasets for benchmarking when available. This problem is inherent to the field of affective computing and will only improve as datasets increase in diversity and representation and preprocessing pipelines improve (e.g., faces with darker pigmentation are often more difficult to detect) [91,92]. Fourth, our toolbox currently only includes detection of core facial features (i.e., facial landmarks, action units, and emotions) but there are additional signals in the face that can be informative for social science researchers. Head pose can be used to detect nodding or a shaking of the head which can be signals of consent or dissent in social interactions. Gaze extracted from face videos can be used to infer the attention of the recorded individual. Heart rate and respiration can also be extracted from face videos [93] which can be used to infer arousal or stress levels of the recorded individual. Models for detecting these facial features could be implemented in future versions of Py-Feat pending community interest.

The modular architecture of the Py-feat toolbox should theoretically be able to flexibly accommodate future developments in facial expression research. For example, adding improved models for our existing detection suite should be relatively straightforward assuming the models are trained using pytorch. New functionality can easily be added to the detector class in the form of a new method. Finally, new types of data can be accommodated by adding a new data loader class and data type specific models. For example, as 3D faces using depth cameras or thermal cameras become more ubiquitous accompanying rapid developments in virtual and augmented reality research, researchers can train new models to detect facial expression features, which can be incorporated into the toolbox without impacting extant functionality. We also hope that the research community will contribute new tutorials to our documentation to accelerate the pace of discovery in the field.

In summary, we introduce Py-Feat, an open source full stack framework implemented in Python for performing facial expression analysis from detection, preprocessing, analysis, and visualization. This work leverages efforts from the broader affective computing community by relying on high quality datasets, state of the art models, and building on other open source efforts such as OpenFace. We hope others in the community may be interested in improving this toolbox by providing feedback and bug reports, and also contributing bug fixes, new models and features. We have outlined our contribution guidelines as well as the necessary code and tutorials on how to replicate our work on our main project website (https://py-feat.org). We look forward to the increasing synergy between the fields of computer science and social science and welcome feedback and suggestions from the broader community as we continue to refine and add features to the Py-Feat platform.

PYTHON FACIAL EXPRESSION ANALYSIS TOOLBOX

## Supplementary Materials

*Pre-trained Facial Detectors*

The `Detector` module offers several pre-trained models for detecting each of the following facial features: (a) finding a face in an image or video frame ("face-model"), (b) locating facial landmarks ("landmark model"), (c) detecting activations of facial muscle action units ("AU model"), and (d) detecting displays of canonical emotional expressions ("emotion model"). These models are designed to be modular so users can decide which algorithms to use for each detection task based on their needs for accuracy and speed. In general, we included models with high reported accuracy, written in Python, easy to install (e.g., Pytorch [57] for neural network models and scikit-learn [94] for statistical models), and open to use for academic research. We have trained several models specifically for Py-Feat and describe the training procedures in detail here.

AU Detection

Py-Feat includes two AU detectors which were based on the robust model included in OpenFace outlined in Baltrusaitis et al. (2015) [95]. Following face and landmark detection, we used Histogram of Oriented Gradients (HOGs) as features in predicting action unit activations. HOGs are feature descriptors that describe an image as a distribution of orientations such as edges and corners measured across the image and have been proven effective in identifying people in images as well as action units [95,96]. We first preprocessed each image by aligning the detected faces using the interocular distance to a neutral facial expression. We then detected the facial landmarks for the aligned faces and applied a convex hull to mask out the background irrelevant to the face. To include facial features of the forehead, a convex hull was applied with the eyebrows shifted upwards 1.5 times the distance between the eyebrows and the upper eye landmarks. We extracted HOGs using the scikit-image implementation [97] with 8 orientations, 8x8 pixels per cell, and 2x2 cells per block which led to a total of 5,408 HOG features. We then applied a principal component analysis (PCA) to retain 95% of the variance, which compressed the dimensionality of these features down to 1,195 while also removing noise. The PCA reduced HOG features were then used to predict individual action units using two statistical learning algorithms, specifically a linear Support Vector Machine classifier [76] implemented in scikit-learn [55] (Feat-SVM) and an XGBoost classifier (Feat-XGB)[77]. Both models were trained using multiple publicly available datasets including BP4D [32], BP4D+, DISFA [31], CK+ [30], Shoulder Pain [98] and [99–101]. Aggregating across these datasets enabled us to make predictions about a larger number of AUs (20 in total) and expose our model to both controlled and in-the-wild data. Hyperparameters were tuned with a grid search during training using 3-fold cross validation. Model performance was evaluated using F1 scores, an accuracy metric for binary classification, defined as:

$$F1 = 2 * \left( \frac{precision * recall}{precision + recall} \right)$$
(eq1)

where precision is the number of true positives divided by the total number of positive results:

$$precision = \frac{true\ positive}{true\ positive + false\ positive} \quad \text{(eq2)}$$

and recall is the proportion of true positives relative to the ground truth:

$$recall = \frac{true\ positive}{true\ positive + false\ negative} \quad \text{(eq3)}$$

F1 scores range from 0 to a perfect precision and recall of 1.0.

Emotion detectors

Emotion detectors are trained on manually posed or naturalistically elicited emotional facial expressions which allows detectors to classify new images based on how much a face resembles a canonical emotional facial expression. Py-feat also includes two emotion detectors. The Residual Masking Network (ResMaskNet) [83] is an end-to-end convolutional neural network model that combines deep residual networks with masking blocks. The masking blocks help focus the model's attention on local regions of interest to refine its feature map for more fine-grained predictions and the residual structure helps to maintain performances in deeper layers. ResMaskNet achieved state of the art performance on the facial expression recognition (FER) 2013 [102] dataset at the time of preparing this article. Despite its accuracy, ResMaskNet has a large memory footprint (500MB) due to the depth of the architecture. We also trained an emotion detector model using an identical pipeline as our statistical learning AU models. This includes performing face alignment, applying a convex hull, and extracting HOG features, which are compressed using a PCA model that retains 95% of the variance. These features are used to classify the presence of each categorical emotion category using linear SVM implemented in scikit-learn[55]. The model was trained using the ExpW [103], CK+ [30] and JAFFE [104] facial expressions datasets with a 3-fold cross validation for identifying the best hyperparameters. Similar to AU detectors, we evaluate model performance with F1 scores for each emotion category.

AU Visualization Model

Py-feat includes a model to visualize facial expression results on an anonymized and stylized face. Using this model, users can visualize the action units and their accompanying 2D landmark deformation on a standard face from any combination of action unit activations identified from their analyses. This can be useful for visualizing aspects of a model in an intuitive manner similar to how brain imaging software overlays statistical maps on a canonical brain [22,86]. We trained this action unit to landmark model on 20 action units (AUs 1, 2, 4, 5, 6, 7, 9, 10, 12, 14, 15, 17, 18, 20, 23, 24, 25, 26, 28, 43) with a subset of images from the EmotioNet [105], BP4D[32], and Extended DISFA Plus [33] datasets to balance the representation of each AU. We chose these datasets because they have both ground truth Action Unit labels. We used our toolbox with the Feat-RetinaFace face detector and MobileNets landmark detector to detect the landmarks on these images. We aligned these landmarks to a neutral face with an affine transformation using the facial landmarks and fit a Partial Least Squares Regression model with 20 components to predict these aligned landmarks from the ground truth action unit labels provided by the datasets using 3-fold cross-validation. Code to reproduce training and testing our visualization model is available in the Py-Feat Training Visualization Model Tutorial.

Overall, the PLS model achieved a cross-validated $r^2$ of 0.155 in predicting landmark coordinate positions on 10,000 sample images. We used our model to illustrate how visualizations can be created in two ways. First we visualize *emotions* by detecting happy, sad, surprise, and anger expressions from single images in the CK+ [30] dataset using the Residual Masking Network implemented in Py-Feat and then passing the AU vectors detected by the Feat-XGB AU classifier to our visualization model (Figure 3A). This is all handled seamlessly using the `detector.plot_detections()`. In principle, Py-Feat's visualization model can generate a face from *any* 20 element array of numerical values between 0 and 1. This enables Py-Feat's second mode of visualization handled by the `plot_face()` and `animate_face()` functions, which can *activate* one or more of AUs and their underlying muscles e.g. AU1 (inner brow raiser), AU12 (lip corner puller), etc (Figure 3B).

## *Datasets*

Training Datasets

**BP4D** [32] is a dataset that includes 8 well-validated emotion induction tasks to elicit multiple emotional expressions (happiness, sadness, surprise, embarrassment, fear, pain, anger and disgust). It contains 41 subjects with 23 female participants and 18 male participants. Participants were recruited from universities, 18-29 years of age, 11 Asians, 6 African American, 4 Hispanic and 20 Euro-American. Two expert FACS coders independently annotated AUs for each frame. A total of 23 videos containing 140,000 frames annotated with binary Action Unit labels (present or not present) and facial landmarks, automatically detected by SDM [106]. We used annotations for AU1, 2, 4, 6,7, 10, 12, 14, 15, 17, 23, and 24 ( on average occurs more than 5% of all the labels)

**BP4D+** [107] contains 140 subjects with 82 females and 58 males. Ten tasks are designed to elicit one of the emotional expressions including (happiness, surprise, sadness, skeptical, embarrassment, fear, pain, anger and disgust). Out of the 10 tasks, 4 tasks, with 197,875 frames have manual Action Unit annotations by expert FACS coders. Participants were recruited at Binghamton University and varied in age (18-66 years old) and ethnicity/race (46 Asians, 15 African American, 14 Hispanic, 64 Euro-American, 1 Others). We used annotations for AU1, 2, 4, 6, 7, 9, 10, 12, 14, 15, 17, 23, and 24 ( on average occurs more than 5% of all the labels).

**DISFA** [108] contains 27 Participants (15 male and 12 female, 18-50 years old, 1 Asian, 1 African American, 2 Hispanic and 21 Euro-American) that watched a 4-minute video clip designed to elicit a certain emotional expression. For each participant, 4,845 video frames were captured and manually annotated by a single expert FACS coder. AU intensity is rated on a six-point ordinal scale from 0 to 5. We binarized AUs using a threshold of 2. We used annotations for AUs 1, 2, 4, 5, 6, 9, 12, 17, 20, 25, and 26.

**DISFA+** [109] is an extended dataset from the original DISFA dataset. It includes 9 participants (4 males and 5 females, 18-50 years old, 1 Asian, 1 African American and 7 Euro-American). DISFA+ contains both posed and spontaneous facial expressions. Participants first watched a 3-minute video clip intended to elicit a certain emotional feeling. In a following experiment, each participant was asked to imitate 30 facial action units, either single AU or combinations of AUs, and 12 facial expressions corresponding to emotions. A trained FACS coder annotated AU intensities (from a ordinal scale of 0 to 5) for a total of over 57,000 frames. We used annotations for AU1,2,4,5,6,9,12,17,20,25,26.

**CK+** [110] contains 593 video sequences from 123 subjects (18-50 years old; 85 females, 38 males; 81% Euro-American, 13% Afro-American, and 6% other groups). Participants were instructed to perform a total of 23 facial displays, including both single AU expressions and combined AU expressions. Trained FACS coders annotated 327 such sequences, and a total number of 1281 images were used. We used annotations for AUs 1, 2, 4, 5, 6, 7, 9, 10, 11, 12, 14, 15, 16, 17, 20, 23, 24, 25, 26, 27, 38, 39, and 43.

**JAFFE** [104] contains ten female Japanese college students. Each participant posed 3 or 4 examples for each of the 6 basic emotion facial expressions plus a neutral face. JAFFE is a relatively small dataset with a total number of 219 images.

**EmotioNet** [101] Contains approximately one million images of facial expressions with Facial Action Unit labels of different gender and ethnicities. The images are downloaded from the Internet. 100,000 images were annotated by trained FACS coders and 900,000 were automatically annotated. The dataset contains faces of different ages, gender, ethnicity and emotional expressions. We used AU annotations for AUs 1, 2, 4, 5, 6, 9, 12, 17, 20, 25, 26, and 43.

**AffectNet** [84] contains 440,000 images collected in the wild downloaded from the Internet with various gender and ethnicity information. Images are manually annotated with eight different emotion categories including: neutral, surprise, happy, fear, sad, disgust, contempt and anger.

**UNBC-McMaster Shoulder Pain** dataset [111] contains 200 face videos from 25 different patients suffering from shoulder pain (total 48,398 frames). Participants were asked to perform a series of either active or passive range-of-motion tests. AUs 4, 6, 7, 9, 10, 12, 20, 25, 26, 27, and 43 were rated on a 5-level intensity by 3 independent certified FACS coders, and a fourth FACS coder reviewed the coding.

Test Datasets

**WIDER FACE** [68] Contains images collected in the wild retrieved from search engines (e.g., Google or Bing). The bounding boxes for each face were manually annotated with a total of 32,203 images with 393,703 labeled faces. This dataset is a standard for benchmarking face detection algorithms in data competitions and includes small, occluded, and upside down faces.

**300W** [112] Contains both in-door and in-the-wild facial images retrieved from google searches. Facial landmarks for each image were semi-automatically annotated by the AOM algorithm [113,114]. The 300W dataset covers a wide variation in luminance, pose, identity, expression, occlusion and face size.

**NAMBA** [85] contains 288 images collected from 12 Japanese participants (6 females and 6 males). Participants were told to imitate certain facial expressions, and a camera videotaped their expressions at angles of 0°, 15°, 30° and 45°. Facial action units (FACS) were annotated for each image by a professional annotator. The annotated AUs include AUs 1, 2, 4, 5, 6, 7, 9, 10, 12, 14, 15, 17, 18, 20, 23, 24, 25, 26, 27, and 43.

**BIWI-Kinect** [75] contains a total number of 15,678 frames collected from 20 subjects (6 females and 14 males) in a controlled in-door environment setting covering a wide range of poses. For each frame, a depth image, the corresponding rgb image, and the head pose annotation is provided. The head pose range covers about +-75 degrees yaw and +-60 degrees pitch.

## Robustness Tests

In addition to our assessing the performance of our detector models on standard benchmark datasets, we were also evaluated the robustness of the detector models included in Py-feat to different types of real world scenarios that are known to create problems for computer vision models including variations in luminance, occlusions of specific regions of the face, and also head rotation. A brief summary of these results are available in Figure 2 for the default models in py-feat. We have also included tables that include the results of our robustness experiments for all detector models included in the toolbox. Table S1 includes results for all face detection models. Table S2 includes results for the landmark detector models. Table S3 includes results for pose estimation models. Table S4 includes results for action unit detectors. Table S5 includes results for the emotion category models. Finally, we include the performance of our action unit detector models in comparison to OpenFace on the Namba head rotation dataset [85].

## Supplementary Tables

**Table S1.** Robustness Test results for face Bounding Box detection with the wider face dataset. Values are Average Precision (AP) for images in each difficulty level (total 3 levels: easy, medium and hard), where higher values indicate better performance. We conducted 5 robustness tests for each algorithm (lower/higher luminance, eyes/nose/mouth masking). Each box indicates the performance of each algorithm on the original test set, and on each robustness test.

| Model | Test | Easy (AP) | Medium (AP) | Hard (AP) |
|---|---|---|---|---|
| *Img2pose constrained* | Baseline | 0.647 | 0.588 | 0.324 |
| | Luminance High | 0.588 | 0.532 | 0.283 |
| | Luminance Low | 0.586 | 0.533 | 0.284 |
| | Mask Eyes | 0.403 | 0.367 | 0.201 |
| | Mask Nose | 0.276 | 0.223 | 0.101 |
| | Mask Mouth | 0.368 | 0.304 | 0.142 |
| *Img2pose unconstrained* | Baseline | 0.856 | 0.814 | 0.574 |
| | Luminance High | 0.838 | 0.786 | 0.527 |
| | Luminance Low | 0.829 | 0.784 | 0.547 |
| | Mask Eyes | 0.703 | 0.680 | 0.488 |
| | Mask Nose | 0.615 | 0.510 | 0.289 |
| | Mask Mouth | 0.733 | 0.633 | 0.357 |
| *Faceboxes* | Baseline | 0.537 | 0.348 | 0.147 |
| | Luminance High | 0.508 | 0.343 | 0.145 |
| | Luminance Low | 0.483 | 0.308 | 0.129 |
| | Mask Eyes | 0.303 | 0.182 | 0.076 |
| | Mask Nose | 0.170 | 0.102 | 0.043 |
| | Mask Mouth | 0.312 | 0.189 | 0.079 |
| *MTCNN* | Baseline | 0.725 | 0.718 | 0.473 |
| | Luminance High | 0.657 | 0.611 | 0.366 |
| | Luminance Low | 0.665 | 0.658 | 0.415 |
| | Mask Eyes | 0.328 | 0.262 | 0.122 |
| | Mask Nose | 0.392 | 0.372 | 0.200 |
| | Mask Mouth | 0.625 | 0.595 | 0.338 |
| *RetinaFace* | Baseline | 0.760 | 0.669 | 0.347 |
| | Luminance High | 0.740 | 0.656 | 0.357 |
| | Luminance Low | 0.732 | 0.632 | 0.320 |
| | Mask Eyes | 0.471 | 0.393 | 0.200 |
| | Mask Nose | 0.317 | 0.223 | 0.104 |
| | Mask Mouth | 0.551 | 0.413 | 0.187 |

**Table S2**: Robustness Test results for Pose detection algorithms with the BIWI-Kinect dataset. Values are Absolute error in degrees for Pitch, Roll and Yaw, where lower values indicate better performance. We conducted 5 robustness tests for each algorithm (lower/higher luminance, eyes/nose/mouth masking). Each box indicates the performance of each algorithm on the original test set, and on each robustness test.

| Model | Test | Pitch MAE | Roll MAE | Yaw MAE | Overall MAE |
|---|---|---|---|---|---|
| *Img2pose constrained* | Baseline | 4.57 | 4.54 | 3.39 | 4.16 |
|  | Luminance High | 4.89 | 4.71 | 3.44 | 4.34 |
|  | Luminance Low | 5.05 | 4.78 | 3.50 | 4.44 |
|  | Mask Eyes | 5.67 | 5.68 | 3.52 | 4.94 |
|  | Mask Nose | 7.57 | 5.30 | 4.14 | 5.67 |
|  | Mask Mouth | 5.87 | 6.33 | 4.29 | 5.50 |
| *Img2pose unconstrained* | Baseline | 6.25 | 4.54 | 3.38 | 4.73 |
|  | Luminance High | 6.42 | 4.79 | 3.38 | 4.87 |
|  | Luminance Low | 6.42 | 4.71 | 3.53 | 4.89 |
|  | Mask Eyes | 5.91 | 4.93 | 3.46 | 4.77 |
|  | Mask Nose | 7.57 | 5.77 | 3.88 | 5.74 |
|  | Mask Mouth | 6.41 | 4.35 | 4.47 | 5.08 |

**Table S3**: Robustness Test results for face landmark detection algorithms with the 300W dataset. Values are normalized mean squared error (nMSE), where lower values indicate better performance. We conducted 5 robustness tests for each algorithm (lower/higher luminance, eyes/nose/mouth masking). Each row shows results for each landmark algorithm in our toolbox, and the columns show each robustness test.

| Model | Baseline | Luminance Low | Luminance High | Mask Mouth | Mask Nose | Mask Eyes |
|---|---|---|---|---|---|---|
| Feat-MobileNet | 5.78 | 7.12 | 6.48 | 15.84 | 19.12 | 8.12 |
| Feat-MobileFaceNet | [4.99] | 5.12 | 5.11 | 7.85 | 9.12 | 6.21 |
| Feat-PFLD | 5.39 | 5.63 | 5.53 | 8.69 | 9.89 | 6.67 |

**Table S4**: Robustness Test results for Action Unit detection algorithms with the DISFA+ dataset. Values are F1 scores for each Action Unit, where higher values indicate better performance. We conducted 5 robustness tests for each algorithm (lower/higher luminance, eyes/nose/mouth masking). Each box indicates the performance of each algorithm on the original test set, and on each robustness test.

| Model | Test | AU1 | AU2 | AU4 | AU5 | AU6 | AU9 | AU12 | AU15 | AU17 | AU20 | AU25 | AU26 | Average |
|---|---|---|---|---|---|---|---|---|---|---|---|---|---|---|
| Feat-XGB | Baseline | [.60] | [.54] | [.64] | [.**57**] | [.71] | [.39] | [.78] | [.25] | [.26] | [.35] | [.84] | [.56] | [.54] |
| | Luminance High | .56 | .58 | .39 | .45 | .60 | .27 | .72 | .28 | .24 | .25 | .69 | .52 | .46 |
| | Luminance Low | .60 | .56 | .55 | .54 | .60 | .34 | .73 | .26 | .25 | .26 | .69 | .53 | .49 |
| | Mask Eyes | .41 | .35 | .43 | .16 | .57 | .22 | .71 | .30 | .25 | .28 | .70 | .62 | .42 |
| | Mask Nose | .57 | .56 | .41 | .51 | .68 | .31 | .73 | .27 | .21 | .22 | .69 | .47 | .47 |
| | Mask Mouth | .47 | .47 | .28 | .50 | .60 | .22 | .49 | .11 | .19 | .19 | .44 | .21 | .35 |
| Feat-SVM | Baseline | .50 | .46 | .73 | .46 | .68 | .59 | .78 | .20 | .29 | .19 | .74 | .69 | .53 |
| | Luminance High | .54 | .41 | .45 | .38 | .51 | .47 | .65 | .23 | .30 | .14 | .57 | .34 | .42 |
| | Luminance Low | .46 | .40 | .56 | .45 | .45 | .55 | .66 | .19 | .26 | .13 | .61 | .39 | .43 |
| | Mask Eyes | .37 | .45 | .42 | .35 | .61 | .19 | .76 | .16 | .22 | .21 | .54 | .58 | .41 |
| | Mask Nose | .41 | .41 | .52 | .54 | .64 | .47 | .46 | .15 | .29 | .23 | .48 | .37 | .41 |
| | Mask Mouth | .40 | .40 | .49 | .44 | .24 | .51 | .27 | .13 | .19 | .15 | .43 | .19 | .32 |

**Table S5**. Robustness Test results for Emotion detection algorithms with the subset AffectNet dataset. Values are F1 scores for each Emotion category, where higher values indicate better performance. We conducted 5 robustness tests for each algorithm (lower/higher luminance, eyes/nose/mouth masking). Each box indicates the performance of each algorithm on the original test set, and on each robustness test.

| Model | Test | Anger | Disgust | Fear | Happy | Sad | Surprise | Neutral | Average |
|---|---|---|---|---|---|---|---|---|---|
| *Residual Masking Network* | Baseline | [.53] | [.53] | [.48] | [.77] | [.54] | [.55] | [.49] | [.55] |
| | Luminance High | .49 | .40 | .47 | .77 | .48 | .51 | .46 | .51 |
| | Luminance Low | .44 | .43 | .54 | .76 | .52 | .44 | .43 | .51 |
| | Mask Eyes | .18 | .47 | .15 | .73 | .05 | .25 | .42 | .32 |
| | Mask Nose | .32 | .33 | .26 | .70 | .14 | .53 | .42 | .39 |
| | Mask Mouth | .40 | .14 | .39 | .26 | .48 | .50 | .28 | .35 |
| *Feat-SVM* | Baseline | .39 | .28 | .37 | .64 | .33 | .39 | .34 | .39 |
| | Luminance High | .12 | .04 | .08 | .26 | .04 | .21 | .26 | .14 |
| | Luminance Low | .09 | .04 | .07 | .25 | .07 | .19 | .27 | .14 |
| | Mask Eyes | .12 | .04 | .15 | .24 | .09 | .19 | .27 | .16 |
| | Mask Nose | .25 | .16 | .27 | .16 | .12 | .25 | .26 | .21 |
| | Mask Mouth | .20 | .05 | .03 | .26 | .02 | .17 | .25 | .14 |

# Supplementary Figures

**Figure S1**. Overview of all modules in the Py-Feat toolbox made using the [Github Next Repo Visualization Tool](#)

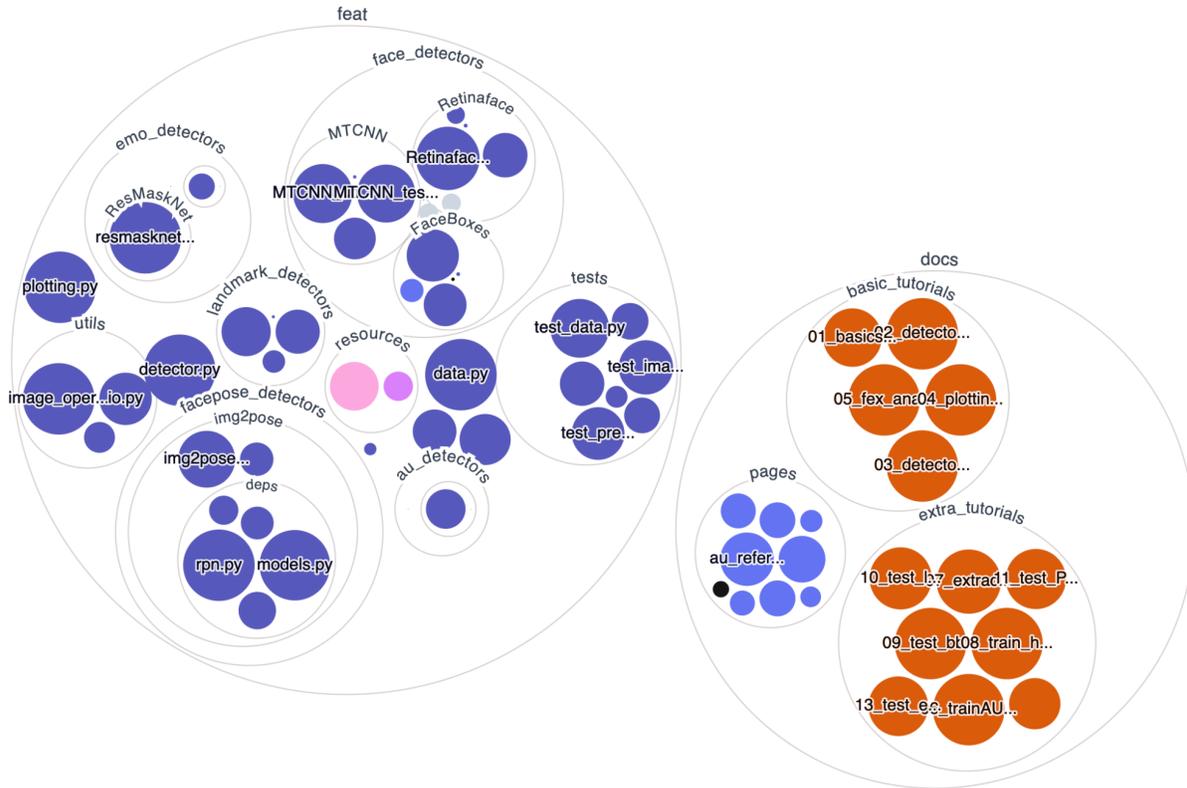

**Figure S2**. Robustness Test results for Action Unit detection algorithms on the Namba head rotation dataset. Head rotation values range from 0° (head on) to 45° rotations. Values are F1 scores for each action unit, where higher values indicate better performance. Each bar indicates the performance of each algorithm on varying degrees of rotation.

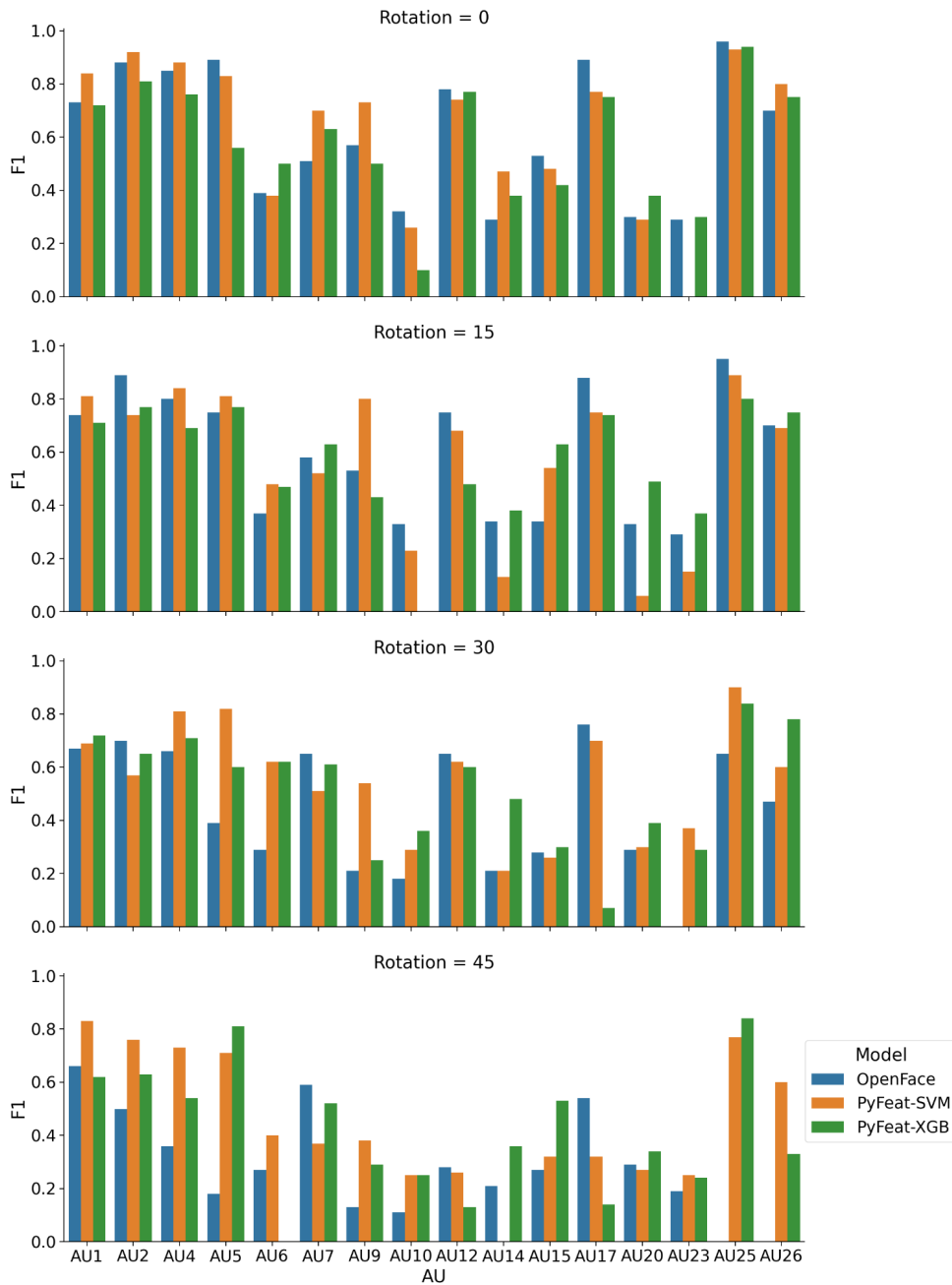

# Code availability

All the code and data to reproduce the results are available at [https://py-feat.org](https://py-feat.org).

# Acknowledgments

The authors would like to acknowledge Filippo Rossi and Nathaniel Hanes for early discussions of this work. We thank Mark Thornton, Emma Templeton, and Wasita Mahaphanit for providing feedback on earlier drafts of this paper. We thank Shushi Namba, Wataru Sato and Sakiko Yoshikawa for generously sharing their data with us. This work was supported by funding from the National Institute of Mental Health R01MH116026, R56MH080716 and the National Science Foundation CAREER 1848370.

# Conflict of interest

The authors declare no competing interests.

# References

1. Darwin, C. *The Expression of the Emotions in Man and Animals*. (1886).
2. Ekman, P. Facial expression and emotion. *Am. Psychol.* **48**, 384 (1993).
3. Ekman, P. & Friesen, W. Facial action coding system: a technique for the measurement of facial movement. *Palo Alto: Consulting Psychologists* (1978).
4. Sayette, M. A., Cohn, J. F., Wertz, J. M. & Perrott, M. A. A psychometric evaluation of the facial action coding system for assessing spontaneous expression. *J. Nonverbal Behav.* (2001).
5. Cohn, J. F., Ambadar, Z. & Ekman, P. Observer-based measurement of facial expression with the Facial Action Coding System. *The handbook of emotion elicitation and assessment* 203–221 (2007).
6. Kilbride, J. E. & Yarczower, M. Ethnic bias in the recognition of facial expressions. *J. Nonverbal Behav.* **8**, 27–41 (1983).
7. Graesser, A. C. *et al.* Detection of emotions during learning with AutoTutor. in *Proceedings of the 28th annual meetings of the cognitive science society* 285–290 (Citeseer, 2006).
8. Fridlund, A. J., Schwartz, G. E. & Fowler, S. C. Pattern recognition of self-reported emotional state from multiple-site facial EMG activity during affective imagery. *Psychophysiology* **21**, 622–637 (1984).
9. Larsen, J. T., Norris, C. J. & Cacioppo, J. T. Effects of positive and negative affect on electromyographic activity over zygomaticus major and corrugator supercilii. *Psychophysiology* **40**, 776–785 (2003).
10. Sayette, M. A. *et al.* Alcohol and group formation: a multimodal investigation of the effects of alcohol on emotion and social bonding. *Psychol. Sci.* **23**, 869–878 (2012).
11. Navarathna, R. *et al.* Predicting movie ratings from audience behaviors. in *IEEE Winter Conference on Applications of Computer Vision* 1058–1065 (2014).
12. Golland, Y., Mevorach, D. & Levit-Binnun, N. Affiliative zygomatic synchrony in co-present strangers. *Scientific Reports* vol. 9 Preprint at https://doi.org/10.1038/s41598-019-40060-4 (2019).
13. Cheong, J. H., Brooks, S. & Chang, L. J. FaceSync: Open source framework for recording facial expressions with head-mounted cameras. *F1000Res.* (2019).
14. Cheong, J. H., Molani, Z., Sadhukha, S. & Chang, L. J. Synchronized affect in shared experiences strengthens social connection. (2020) doi:10.31234/osf.io/bd9wn.
15. De la Torre, F. *et al.* IntraFace. *IEEE Int Conf Autom Face Gesture Recognit Workshops* **1**, (2015).
16. Vemulapalli, R. & Agarwala, A. A compact embedding for facial expression similarity. in *Proceedings of the IEEE/CVF Conference on Computer Vision and Pattern Recognition* 5683–5692 (openaccess.thecvf.com, 2019).
17. Stöckli, S., Schulte-Mecklenbeck, M., Borer, S. & Samson, A. C. Facial expression analysis with AFFDEX and FACET: A validation study. *Behav. Res. Methods* **50**, 1446–1460 (2018).
18. Haines, N., Southward, M. W., Cheavens, J. S., Beauchaine, T. & Ahn, W.-Y. Using computer-vision and machine learning to automate facial coding of positive and negative affect intensity. *PLoS One* **14**, e0211735 (2019).
19. Höfling, T. T. A., Gerdes, A. B. M., Föhl, U. & Alpers, G. W. Read My Face: Automatic Facial Coding Versus Psychophysiological Indicators of Emotional Valence and Arousal. *Front. Psychol.* **11**, 1388 (2020).
20. Dupré, D., Krumhuber, E. G., Küster, D. & McKeown, G. J. A performance comparison of eight commercially available automatic classifiers for facial affect recognition. *PLoS One* **15**, e0231968 (2020).
21. Werner, P. *et al.* Automatic Pain Assessment with Facial Activity Descriptors. *IEEE*

*Transactions on Affective Computing* **8**, 286–299 (2017).
22. Chen, P.-H. A. *et al.* Socially transmitted placebo effects. *Nat Hum Behav* **3**, 1295–1305 (2019).
23. Littlewort, G. C., Bartlett, M. S. & Lee, K. Automatic coding of facial expressions displayed during posed and genuine pain. *Image Vis. Comput.* **27**, 1797–1803 (2009).
24. Wang, Y. *et al.* Automatic Depression Detection via Facial Expressions Using Multiple Instance Learning. in *2020 IEEE 17th International Symposium on Biomedical Imaging (ISBI)* 1933–1936 (2020).
25. Penton-Voak, I. S., Pound, N., Little, A. C. & Perrett, D. I. Personality Judgments from Natural and Composite Facial Images: More Evidence For A 'Kernel Of Truth' In Social Perception. *Soc. Cogn.* **24**, 607–640 (2006).
26. Segalin, C. *et al.* What your Facebook Profile Picture Reveals about your Personality. *Proceedings of the 25th ACM international conference on Multimedia* Preprint at https://doi.org/10.1145/3123266.3123331 (2017).
27. Kachur, A., Osin, E., Davydov, D., Shutilov, K. & Novokshonov, A. Assessing the Big Five personality traits using real-life static facial images. *Sci. Rep.* **10**, 8487 (2020).
28. Kosinski, M. Facial recognition technology can expose political orientation from naturalistic facial images. *Sci. Rep.* **11**, 100 (2021).
29. Kanade, T., Cohn, J. F. & Yingli Tian. Comprehensive database for facial expression analysis. in *Proceedings Fourth IEEE International Conference on Automatic Face and Gesture Recognition (Cat. No. PR00580)* 46–53 (2000).
30. Lucey, P. *et al.* The Extended Cohn-Kanade Dataset (CK ): A complete dataset for action unit and emotion-specified expression. *2010 IEEE Computer Society Conference on Computer Vision and Pattern Recognition - Workshops* Preprint at https://doi.org/10.1109/cvprw.2010.5543262 (2010).
31. Mavadati, S. M., Mahoor, M. H., Bartlett, K., Trinh, P. & Cohn, J. F. DISFA: A Spontaneous Facial Action Intensity Database. *IEEE Transactions on Affective Computing* **4**, 151–160 (2013).
32. Zhang, X. *et al.* BP4D-Spontaneous: a high-resolution spontaneous 3D dynamic facial expression database. *Image Vis. Comput.* **32**, 692–706 (2014).
33. Mavadati, M., Sanger, P. & Mahoor, M. H. Extended disfa dataset: Investigating posed and spontaneous facial expressions. in *proceedings of the IEEE conference on computer vision and pattern recognition workshops* 1–8 (2016).
34. Zhang, Z. *et al.* Multimodal spontaneous emotion corpus for human behavior analysis. in *Proceedings of the IEEE Conference on Computer Vision and Pattern Recognition* 3438–3446 (2016).
35. Krumhuber, E. G., Skora, L., Küster, D. & Fou, L. A Review of Dynamic Datasets for Facial Expression Research. *Emot. Rev.* **9**, 280–292 (2017).
36. Dhall, A., Goecke, R., Joshi, J., Sikka, K. & Gedeon, T. Emotion Recognition In The Wild Challenge 2014: Baseline, Data and Protocol. in *Proceedings of the 16th International Conference on Multimodal Interaction* 461–466 (Association for Computing Machinery, 2014).
37. Qi, P., Zhang, Y., Zhang, Y., Bolton, J. & Manning, C. D. Stanza: A Python Natural Language Processing Toolkit for Many Human Languages. *arXiv [cs.CL]* (2020).
38. Brockman, G. *et al.* OpenAI Gym. *arXiv [cs.LG]* (2016).
39. *iMotions Biometric Research Platform 6.0*. (iMotions A/S, Copenhagen, Denmark, 2016).
40. Van Kuilenburg, H., Den Uyl, M. J., Israël, M. L. & Ivan, P. Advances in face and gesture analysis. *Measuring Behavior 2008* **371**, (2008).
41. Yitzhak, N. *et al.* Gently does it: Humans outperform a software classifier in recognizing subtle, nonstereotypical facial expressions. *Emotion* **17**, 1187–1198 (2017).
42. Krumhuber, E. G., Küster, D., Namba, S. & Skora, L. Human and machine validation of 14

databases of dynamic facial expressions. *Behav. Res. Methods* (2020) doi:10.3758/s13428-020-01443-y.
43. Krumhuber, E. G., Küster, D., Namba, S., Shah, D. & Calvo, M. G. Emotion recognition from posed and spontaneous dynamic expressions: Human observers versus machine analysis. *Emotion* **21**, 447–451 (2021).
44. Littlewort, G. *et al.* The computer expression recognition toolbox (CERT). in *2011 IEEE International Conference on Automatic Face Gesture Recognition (FG)* 298–305 (ieeexplore.ieee.org, 2011).
45. McDuff, D. *et al.* AFFDEX SDK: A Cross-Platform Real-Time Multi-Face Expression Recognition Toolkit. in *Proceedings of the 2016 CHI Conference Extended Abstracts on Human Factors in Computing Systems* 3723–3726 (Association for Computing Machinery, 2016).
46. Baltrusaitis, T., Zadeh, A., Lim, Y. C. & Morency, L. OpenFace 2.0: Facial Behavior Analysis Toolkit. in *2018 13th IEEE International Conference on Automatic Face Gesture Recognition (FG 2018)* 59–66 (2018).
47. Jenkinson, M., Beckmann, C. F., Behrens, T. E. J., Woolrich, M. W. & Smith, S. M. FSL. *Neuroimage* **62**, 782–790 (2012).
48. Cox, R. W. AFNI: software for analysis and visualization of functional magnetic resonance neuroimages. *Comput. Biomed. Res.* **29**, 162–173 (1996).
49. Friston, K. J., Frith, C. D., Liddle, P. F. & Frackowiak, R. S. Comparing functional (PET) images: the assessment of significant change. *J. Cereb. Blood Flow Metab.* **11**, 690–699 (1991).
50. Abraham, A. *et al.* Machine learning for neuroimaging with scikit-learn. *Front. Neuroinform.* **8**, 14 (2014).
51. Ekman, P. & Rosenberg, E. L. *What the Face Reveals: Basic and Applied Studies of Spontaneous Expression Using the Facial Action Coding System (FACS)*. (Oxford University Press, 1997).
52. Harris, C. R. *et al.* Array programming with NumPy. *Nature* **585**, 357–362 (2020).
53. McKinney, W. & Others. pandas: a foundational Python library for data analysis and statistics. *Python for High Performance and Scientific Computing* **14**, 1–9 (2011).
54. Jones, E., Oliphant, T. & Peterson, P. {SciPy}: Open source scientific tools for {Python}. Preprint at http://www.scipy.org (2001--).
55. Pedregosa, F. *et al.* Scikit-learn: Machine Learning in Python. *J. Mach. Learn. Res.* **12**, 2825–2830 (2011).
56. Abadi, M. *et al.* Tensorflow: A system for large-scale machine learning. in *12th ${USENIX} symposium on operating systems design and implementation ({OSDI}$ 16)* 265–283 (2016).
57. Paszke, A. *et al.* PyTorch: An Imperative Style, High-Performance Deep Learning Library. *arXiv [cs.LG]* (2019).
58. Hunter. Matplotlib: A 2D Graphics Environment. **9**, 90–95 (2007).
59. Waskom, M. seaborn: statistical data visualization. *J. Open Source Softw.* **6**, 3021 (2021).
60. Chang, L. J. *et al. ljchang/dartbrains: An online open access resource for learning functional neuroimaging analysis methods in Python*. (2020). doi:10.5281/zenodo.3909718.
61. Chang, L. *et al. naturalistic-data-analysis/naturalistic_data_analysis: Version 1.0*. (2020). doi:10.5281/zenodo.3937849.
62. Executable Books Community. *Jupyter Book*. (2020). doi:10.5281/zenodo.4539666.
63. Zhang, S. *et al.* FaceBoxes: A CPU real-time face detector with high accuracy. in *2017 IEEE International Joint Conference on Biometrics (IJCB)* 1–9 (2017).
64. Zhang, L. *et al.* Multi-Task Cascaded Convolutional Networks Based Intelligent Fruit Detection for Designing Automated Robot. *IEEE Access* **7**, 56028–56038 (2019).
65. Zhang, N., Luo, J. & Gao, W. Research on Face Detection Technology Based on MTCNN.

PYTHON FACIAL EXPRESSION ANALYSIS TOOLBOX

in *2020 International Conference on Computer Network, Electronic and Automation (ICCNEA)* 154–158 (2020).
66. Deng, J. *et al.* RetinaFace: Single-stage Dense Face Localisation in the Wild. *arXiv [cs.CV]* (2019).
67. Yang, S., Luo, P., Loy, C.-C. & Tang, X. Wider face: A face detection benchmark. in *Proceedings of the IEEE conference on computer vision and pattern recognition* 5525–5533 (2016).
68. Yang, S., Luo, P., Loy, C. C. & Tang, X. WIDER FACE: A Face Detection Benchmark. *arXiv [cs.CV]* (2015).
69. Shen, J. *et al.* The first facial landmark tracking in-the-wild challenge: Benchmark and results. in *Proceedings of the IEEE international conference on computer vision workshops* 50–58 (2015).
70. Sagonas, C., Antonakos, E., Tzimiropoulos, G., Zafeiriou, S. & Pantic, M. 300 Faces In-The-Wild Challenge: database and results. *Image Vis. Comput.* **47**, 3–18 (2016).
71. Guo, X. *et al.* PFLD: A Practical Facial Landmark Detector. *arXiv [cs.CV]* (2019).
72. Howard, A. G. *et al.* MobileNets: Efficient Convolutional Neural Networks for Mobile Vision Applications. *arXiv [cs.CV]* (2017).
73. Chen, S., Liu, Y., Gao, X. & Han, Z. MobileFaceNets: Efficient CNNs for Accurate Real-Time Face Verification on Mobile Devices. *arXiv [cs.CV]* (2018).
74. Sagonas, C., Tzimiropoulos, G., Zafeiriou, S. & Pantic, M. A semi-automatic methodology for facial landmark annotation. in *Proceedings of the IEEE conference on computer vision and pattern recognition workshops* 896–903 (2013).
75. Fanelli, G., Dantone, M., Gall, J., Fossati, A. & Van Gool, L. Random Forests for Real Time 3D Face Analysis. *Int. J. Comput. Vis.* **101**, 437–458 (2013).
76. Chang, C.-C. & Lin, C.-J. LIBSVM: A library for support vector machines. *ACM Trans. Intell. Syst. Technol.* **2**, 1–27 (2011).
77. Chen, T. & Guestrin, C. XGBoost: A Scalable Tree Boosting System. in *Proceedings of the 22nd ACM SIGKDD International Conference on Knowledge Discovery and Data Mining* 785–794 (Association for Computing Machinery, 2016).
78. Jack, R. E., Garrod, O. G. B., Yu, H., Caldara, R. & Schyns, P. G. Facial expressions of emotion are not culturally universal. *Proc. Natl. Acad. Sci. U. S. A.* **109**, 7241–7244 (2012).
79. Cowen, A. S. *et al.* Sixteen facial expressions occur in similar contexts worldwide. *Nature* **589**, 251–257 (2021).
80. Barrett, L. F., Adolphs, R., Marsella, S., Martinez, A. M. & Pollak, S. D. Emotional Expressions Reconsidered: Challenges to Inferring Emotion From Human Facial Movements. *Psychological Science in the Public Interest* vol. 20 1–68 Preprint at https://doi.org/10.1177/1529100619832930 (2019).
81. Saxe, R. & Houlihan, S. D. Formalizing emotion concepts within a Bayesian model of theory of mind. *Curr Opin Psychol* **17**, 15–21 (2017).
82. Haines, N. *et al.* Regret Induces Rapid Learning from Experience-based Decisions: A Model-based Facial Expression Analysis Approach. *bioRxiv* 560011 (2019) doi:10.1101/560011.
83. Luan, P., Huynh, V. & Tuan Anh, T. Facial Expression Recognition using Residual Masking Network. in *IEEE 25th International Conference on Pattern Recognition* 4513–4519 (2020).
84. Mollahosseini, A., Hasani, B. & Mahoor, M. H. AffectNet: A database for facial expression, valence, and arousal computing in the wild. *IEEE Trans. Affect. Comput.* **10**, 18–31 (2019).
85. Namba, S., Sato, W. & Yoshikawa, S. Viewpoint Robustness of Automated Facial Action Unit Detection Systems. *NATO Adv. Sci. Inst. Ser. E Appl. Sci.* **11**, 11171 (2021).
86. Chang, L. J. *et al.* Endogenous variation in ventromedial prefrontal cortex state dynamics during naturalistic viewing reflects affective experience. *Sci Adv* **7**, (2021).
87. Watson, D. M., Brown, B. B. & Johnston, A. A data-driven characterisation of natural facial

expressions when giving good and bad news. *PLoS Comput. Biol.* **16**, e1008335 (2020).
88. Worsley, K. J. & Friston, K. J. Analysis of fMRI Time-Series Revisited—Again. *Neuroimage* **2**, 173–181 (1995).
89. Chang, L. J. *et al.* Endogenous variation in ventromedial prefrontal cortex state dynamics during naturalistic viewing reflects affective experience. *bioRxiv* 487892 (2018) doi:10.1101/487892.
90. Jack, R. E., Garrod, O. G. B. & Schyns, P. G. Dynamic facial expressions of emotion transmit an evolving hierarchy of signals over time. *Curr. Biol.* **24**, 187–192 (2014).
91. Rhue, L. Racial Influence on Automated Perceptions of Emotions. (2018) doi:10.2139/ssrn.3281765.
92. Nagpal, S., Singh, M., Singh, R. & Vatsa, M. Deep Learning for Face Recognition: Pride or Prejudiced? *arXiv [cs.CV]* (2019).
93. McDuff, D., Gontarek, S. & Picard, R. Remote measurement of cognitive stress via heart rate variability. *Conf. Proc. IEEE Eng. Med. Biol. Soc.* **2014**, 2957–2960 (2014).
94. Pedregosa, F. *et al.* Scikit-learn: Machine Learning in Python. *J. Mach. Learn. Res.* **12**, 2825–2830 (2011).
95. Baltrušaitis, T., Mahmoud, M. & Robinson, P. Cross-dataset learning and person-specific normalisation for automatic Action Unit detection. in *2015 11th IEEE International Conference and Workshops on Automatic Face and Gesture Recognition (FG)* vol. 06 1–6 (2015).
96. Dalal, N. & Triggs, B. Histograms of oriented gradients for human detection. in *2005 IEEE Computer Society Conference on Computer Vision and Pattern Recognition (CVPR'05)* vol. 1 886–893 vol. 1 (2005).
97. van der Walt, S. *et al.* scikit-image: image processing in Python. *PeerJ* **2**, e453 (2014).
98. Lucey, P., Cohn, J. F., Prkachin, K. M., Solomon, P. E. & Matthews, I. Painful data: The UNBC-McMaster shoulder pain expression archive database. in *2011 IEEE International Conference on Automatic Face Gesture Recognition (FG)* 57–64 (2011).
99. Fabian Benitez-Quiroz, C., Srinivasan, R., Feng, Q., Wang, Y. & Martinez, A. M. EmotioNet Challenge: Recognition of facial expressions of emotion in the wild. *arXiv [cs.CV]* (2017).
100. Benitez-Quiroz, C. F., Wang, Y. & Martinez, A. M. Recognition of Action Units in the Wild with Deep Nets and a New Global-Local Loss. in *2017 IEEE International Conference on Computer Vision (ICCV)* 3990–3999 (2017).
101. Benitez-Quiroz, C. F., Srinivasan, R. & Martinez, A. M. EmotioNet: An accurate, real-time algorithm for the automatic annotation of a million facial expressions in the wild. in *2016 IEEE Conference on Computer Vision and Pattern Recognition (CVPR)* 5562–5570 (IEEE, 2016).
102. Goodfellow, I. J. *et al.* Challenges in representation learning: A report on three machine learning contests. *Neural Networks* vol. 64 59–63 Preprint at https://doi.org/10.1016/j.neunet.2014.09.005 (2015).
103. Zhang, Z., Luo, P., Loy, C. C. & Tang, X. From facial expression recognition to interpersonal relation prediction. *Int. J. Comput. Vis.* **126**, 550–569 (2018).
104. Lyons, M., Kamachi, M. & Gyoba, J. *The Japanese Female Facial Expression (JAFFE) Dataset*. (1998). doi:10.5281/zenodo.3451524.
105. Fabian Benitez-Quiroz, C., Srinivasan, R. & Martinez, A. M. Emotionet: An accurate, real-time algorithm for the automatic annotation of a million facial expressions in the wild. in *Proceedings of the IEEE conference on computer vision and pattern recognition* 5562–5570 (2016).
106. Xiong, X. & De la Torre, F. Supervised Descent Method and Its Applications to Face Alignment. in *2013 IEEE Conference on Computer Vision and Pattern Recognition* 532–539 (2013).
107. Zhang, Z. *et al.* Multimodal Spontaneous Emotion Corpus for Human Behavior Analysis. in

*2016 IEEE Conference on Computer Vision and Pattern Recognition (CVPR)* 3438–3446 (2016).
108. Mavadati, S. M., Mahoor, M. H., Bartlett, K., Trinh, P. & Cohn, J. F. DISFA: A spontaneous facial action intensity database. *IEEE Trans. Affect. Comput.* **4**, 151–160 (2013).
109. Mavadati, M., Sanger, P. & Mahoor, M. H. Extended DISFA Dataset: Investigating Posed and Spontaneous Facial Expressions. in *2016 IEEE Conference on Computer Vision and Pattern Recognition Workshops (CVPRW)* 1452–1459 (2016).
110. Lucey, P. *et al.* The Extended Cohn-Kanade Dataset (CK+): A complete dataset for action unit and emotion-specified expression. in *2010 IEEE Computer Society Conference on Computer Vision and Pattern Recognition - Workshops* (IEEE, 2010). doi:10.1109/cvprw.2010.5543262.
111. Lucey, P., Cohn, J. F., Prkachin, K. M., Solomon, P. E. & Matthews, I. Painful data: The UNBC-McMaster shoulder pain expression archive database. in *2011 IEEE International Conference on Automatic Face & Gesture Recognition (FG)* 57–64 (2011).
112. Sagonas, C., Tzimiropoulos, G., Zafeiriou, S. & Pantic, M. 300 faces in-the-wild challenge: The first facial landmark localization challenge. in *Proceedings of the IEEE international conference on computer vision workshops* 397–403 (cv-foundation.org, 2013).
113. Tzimiropoulos, G., Alabort-i-Medina, J., Zafeiriou, S. & Pantic, M. Generic Active Appearance Models Revisited. in *Computer Vision – ACCV 2012* 650–663 (Springer Berlin Heidelberg, 2013).
114. Tzimiropoulos, G., Alabort-i-Medina, J., Zafeiriou, S. P. & Pantic, M. Active Orientation Models for Face Alignment In-the-Wild. *IEEE Trans. Inf. Forensics Secur.* **9**, 2024–2034 (2014).